# Perceptual Distortions and Autonomous Representation Learning in a Minimal Robotic System

## Authors

David Warutumo, Ciira wa Maina

Center for Data Science and Artificial Intelligence, Dedan Kimathi University of Agriculture and Technology

davidwarchy@gmail.com

## Abstract

Autonomous agents, particularly in the field of robotics, rely on sensory information to perceive and navigate their environment. However, these sensory inputs are often imperfect, leading to distortions in the agent's internal representation of the world. This paper investigates the nature of these perceptual distortions and how they influence autonomous representation learning using a minimal robotic system. We utilize a simulated two-wheeled robot equipped with distance sensors and a compass, operating within a simple square environment. Through analysis of the robot's sensor data during random exploration, we demonstrate how a distorted perceptual space emerges. Despite these distortions, we identify emergent structures within the perceptual space that correlate with the physical environment, revealing how the robot autonomously learns a structured representation for navigation without explicit spatial information. This work contributes to the understanding of embodied cognition, minimal agency, and the role of perception in self-generated navigation strategies in artificial life.

## Introduction

The field of Artificial Life (ALife) endeavors to unravel the essential principles underlying life itself through the creation and study of artificial systems exhibiting life-like characteristics. A core pursuit within ALife revolves around understanding how autonomous agents, be they biological or artificial, develop internal representations of their environment. These representations are not mere static maps, but dynamic, evolving constructs that agents use for navigation, interaction with their surroundings, and ultimately, survival [1, 2]. This process of representation learning, a cornerstone of intelligence and autonomy, becomes particularly captivating when we consider agents with limited sensory capabilities, forced to grapple with noisy, incomplete, or distorted information about the world. This scenario, far from being an abstract problem, reflects the reality faced by many biological organisms and presents a significant challenge for the design of robust and adaptable artificial agents.

Traditional approaches to robot navigation often rely on a suite of sophisticated sensors, like lidar and GPS, coupled with intricate mapping algorithms such as Simultaneous Localization and Mapping (SLAM) to build detailed and accurate representations of the environment [3, 4]. These methods, while effective in controlled settings, can be computationally demanding, require precise sensor calibration, and often struggle in dynamic or unstructured environments. Moreover, they may not fully capture the embodied and situated nature of perception and action that characterizes biological agents. Biological organisms, even with comparatively simple sensory systems, demonstrate remarkable abilities to navigate complex environments, adapt to changing conditions, and exhibit robust behavior in the face of uncertainty. This raises a fundamental question: how much complexity in sensing and processing is truly necessary for effective navigation and interaction? Can simpler, more biologically inspired approaches yield comparable or even superior performance in certain contexts?

This research takes a minimalist approach to understanding autonomous representation learning by investigating how a resource-constrained agent, equipped with only basic sensors, can autonomously learn to navigate and interact with its environment. We focus on a minimal robotic system: a simulated two-wheeled robot operating within a simplified, yet representative environment—a square arena. The robot's sensory apparatus consists of a set of distance sensors, providing local information about obstacles and boundaries, and a compass sensor (simulated using an Inertial Measurement Unit or IMU), offering information about its orientation or heading. These basic sensors, while providing only a partial and potentially distorted view of the world, are analogous to the sensory systems found in many simple organisms.

Our central aim is to understand how this minimal agent constructs its perceptual space, the internal representation of its environment derived solely from its limited sensory input. We hypothesize that even with these constraints, coherent and functional representations can emerge, enabling the robot to navigate effectively. The key questions we address include:

1. **Perceptual Space Formation:** How does the robot's limited sensory information translate into an internal representation of its environment? How does the arrangement and characteristics of the sensors influence the structure and dimensionality of this perceptual space? Do certain sensor combinations play a more significant role in shaping the robot's understanding of its surroundings?

2. **Distortions and Transformations:** How do the inherent limitations of the sensors, coupled with the robot's movement patterns, introduce distortions into its perceptual space? Do these distortions lead to non-linear transformations of spatial relationships, such as distances and angles, between the physical world and the robot's internal representation?

3. **Emergence of Structure:** Despite these perceptual distortions, can coherent structures still emerge within the robot's internal representation? Do these structures correlate with meaningful features in the physical environment, such as walls, corners, or open spaces? How does the robot leverage these structures for navigation and decision-making?

4. **Autonomous Learning:** How does the robot autonomously learn to utilize its distorted, yet structured perceptual space for navigation? Does this learning process involve the formation of associations between sensorimotor experiences, and how do these associations contribute to the robot's ability to adapt to its environment?

To investigate these questions, we employ a random exploration strategy, allowing the robot to freely navigate the arena while collecting sensor data. This approach ensures that the robot encounters a wide range of sensorimotor experiences, providing a rich dataset for analysis. We then apply a variety of computational techniques, including path visualization, K-Nearest Neighbors analysis, Convex Hull analysis, clustering, and grid transformations, to characterize the robot's perceptual space and reveal the underlying structures that emerge within it.

By studying this minimal robotic system, we aim to gain insights into the fundamental principles of autonomous representation learning and embodied cognition. Our findings have implications not only for the design of more robust and adaptable robots but also for understanding how biological agents, from simple insects to complex animals, perceive and interact with their world. This research contributes to a deeper understanding of the relationship between perception, action, and representation, bridging the gap between minimal agency and more complex forms of intelligence.

## Related Works

This section explores existing research relevant to the study of perceptual distortions and autonomous representation learning in robotics and artificial life. It highlights key themes and contributions from the literature and positions our work within the broader context of embodied cognition, sensorimotor learning, and minimal agency.

**Embodied Cognition and Sensorimotor Learning:**

The concept of embodied cognition emphasizes the crucial role of the body and its interactions with the environment in shaping cognitive processes [1, 2]. This perspective contrasts with traditional views of cognition as purely symbolic computation, highlighting the importance of sensorimotor experience in grounding and structuring internal representations.

- **Pfeifer and Bongard [3]** argue that intelligence is not solely a product of the brain but rather emerges from the dynamic interplay between brain, body, and world. They advocate for a "how the body shapes the way we think" approach, emphasizing the importance of morphology, materials, and sensorimotor dynamics in shaping cognitive abilities. Our work aligns with this perspective by demonstrating how the robot's physical structure and sensorimotor interactions shape its perceptual space and learning process.

- **Wilson [4]** provides a comprehensive overview of embodied cognition, outlining six core views: cognition is situated, time-pressured, subject to bodily constraints, offloaded onto the environment, based on action, and intrinsically social. Our study focuses on the first four views. The robot's cognition is situated within its environment, its actions are time-pressured by the simulation's time steps, it's subject to the constraints of its sensors and actuators, and aspects of navigation are implicitly offloaded onto the environment's structure.

- **Brooks [5]** introduced the concept of "intelligence without representation," arguing that complex behaviors can emerge from simple sensorimotor couplings without the need for explicit internal models. While our work demonstrates a form of representation learning, it aligns with Brooks's philosophy by showing how minimal agents can achieve robust behavior with relatively simple sensory systems and learning mechanisms.

**Sensorimotor Mapping and Perceptual Spaces:**

Several studies have investigated how agents learn to map sensory inputs to motor outputs and construct internal representations of their environment based on sensorimotor experience.

- **Fitzpatrick et al. [6]** studied the development of sensorimotor mappings in simulated robots using evolutionary algorithms. They demonstrated how robots can evolve effective navigation strategies based on simple sensorimotor couplings, highlighting the power of embodiment in generating adaptive behavior. Our work shares a similar focus on sensorimotor learning but delves deeper into the analysis of perceptual distortions and their impact on representation learning.

- **Tani [7]** proposed a framework for understanding sensorimotor coordination based on dynamical systems theory. He argues that internal representations emerge from the self-organization of sensorimotor loops, enabling agents to adapt to changing environmental conditions. This relates directly to our work in terms of implicit learning of an environmental representation from sensorimotor experience and demonstrates how simple exploration can result in complex navigational behavior without explicit planning.

- **Ay et al. [8]** investigated the role of intrinsic motivations in driving sensorimotor learning. They showed that agents can learn complex skills and explore their environment effectively by seeking out novel sensory experiences. While our current study utilizes random exploration, future work could incorporate intrinsic motivations to enhance the learning process.

**Minimal Agency and Autonomous Navigation:**

Research on minimal agency explores how simple agents with limited cognitive capabilities can exhibit autonomous behavior and adapt to their environment.

- **Braitenberg [9]** presented a series of thought experiments involving simple vehicles with light sensors and motors. He demonstrated how seemingly complex behaviors, such as attraction, aversion, and exploration, can arise from simple sensorimotor connections. Our work builds upon this idea by showing how a slightly more complex robot with distance sensors and a compass can learn to navigate a simple environment.

- **Pfeifer et al. [10]** studied the navigational abilities of a minimal robot equipped with a single light sensor and two motors. They demonstrated how the robot could achieve robust wall-following behavior by exploiting the dynamics of its sensorimotor interactions with the environment. This is conceptually analogous to our setup which relies on sixteen distance sensors.

- **Bongard et al. [11]** investigated the role of morphology in shaping the behavior of simple robots. They showed that robots with different body shapes can exhibit distinct navigational patterns, highlighting the importance of physical embodiment in generating adaptive behavior.

**Spatial Representation and Navigation:**

Understanding how agents represent space and navigate within it is a fundamental challenge in robotics and artificial life.

- **O'Keefe and Nadel [12]** proposed the concept of the "cognitive map," a neural representation of the spatial environment that enables animals to navigate and plan routes. While our robot does not possess a hippocampus or a neural network, the emergence of structured clusters in its perceptual space suggests a rudimentary form of spatial representation.

- **Thrun et al. [13]** developed probabilistic robotics, a framework for robot perception and action under uncertainty. They introduced techniques such as SLAM (Simultaneous Localization and Mapping) that enable robots to build maps and localize themselves within those maps. While our work does not involve explicit map building, it explores how a robot can navigate without a precise metric map by relying on sensorimotor experience.

**Perceptual Distortions and Robustness:**

The issue of perceptual distortions and their impact on robot navigation has also been addressed in the literature.

- **Franz and Mallot [14]** studied the effects of sensor noise and perceptual errors on robot navigation. They demonstrated how robots can compensate for these errors by incorporating feedback mechanisms and adapting their control strategies.

- **Steels [15]** investigated the role of self-organization in the development of robust robot controllers. He showed how robots can adapt to changing environmental conditions and sensor malfunctions by evolving decentralized control systems.

**Connecting to Our Work:**

Our research contributes to the existing literature by focusing specifically on the interplay between perceptual distortions and autonomous representation learning in a minimal robotic system. We go beyond simply demonstrating successful navigation and delve deeper into the analysis of the robot's perceptual space. By using techniques like KNN analysis, convex hull comparisons, and clustering, we reveal how the robot's perception transforms spatial relationships and how coherent structures emerge within the distorted sensorimotor data. This detailed analysis of the robot's internal representation provides valuable insights into the mechanisms underlying embodied cognition and minimal agency, bridging the gap between simple sensorimotor behaviors and more complex forms of spatial representation and navigation. Our findings emphasize the importance of considering perceptual distortions when designing and analyzing autonomous agents and suggest that even with limited and imperfect sensory information, agents can develop robust and adaptive behavior through embodied interaction with their environment. Future work will explore the scalability of these principles to more complex environments and investigate how different sensor modalities and learning mechanisms influence the development of perceptual spaces and autonomous navigation strategies.

# Methods

This expanded section provides a more detailed description of the robotic platform, the data collection process, and the analysis techniques used to investigate perceptual distortions and autonomous representation learning.

**2.1 Robotic Platform and Simulated Environment:**

The experimental platform is a simulated two-wheeled differential drive robot within the Webots robotics simulator [21]. This choice allows for precise control over experimental parameters and eliminates real-world complexities like sensor noise and mechanical variations, facilitating a focused study on perceptual distortions.

- **Robot Structure:** The robot consists of a cylindrical body equipped with two independently controlled wheels and sixteen distance sensors (DS). The DS are uniformly distributed around the robot's circumference, providing a 360-degree range of perception. This arrangement is crucial for capturing the spatial characteristics of the environment from various orientations. An Inertial Measurement Unit (IMU) provides the robot's yaw (orientation) relative to the world frame, serving as a compass.

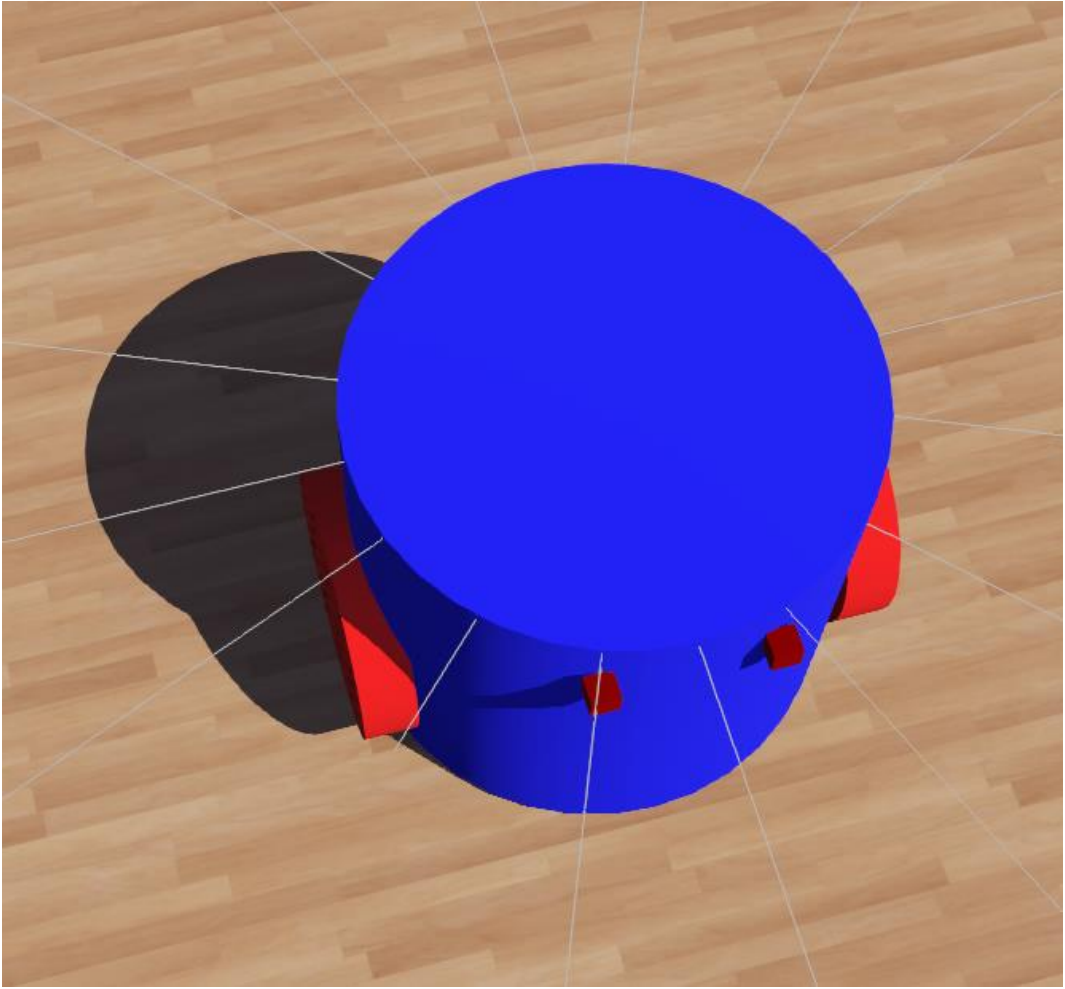

*Figure 1: The Webots [5] robot model with sensor lines shown in grey. The robot has a cylindrical body (blue) and two wheels using differential drive. The eyes are not functional but they are instructive as to the direction of the heading.*

- **Simulated World:** The robot operates within a bounded square arena. The walls of the arena provide boundaries and serve as primary features for the robot to perceive and interact with. This simplified environment allows for a controlled investigation of the relationship between sensor readings and physical location. The dimensions of the arena are 10m x 10m, providing ample space for exploration while maintaining computational efficiency.

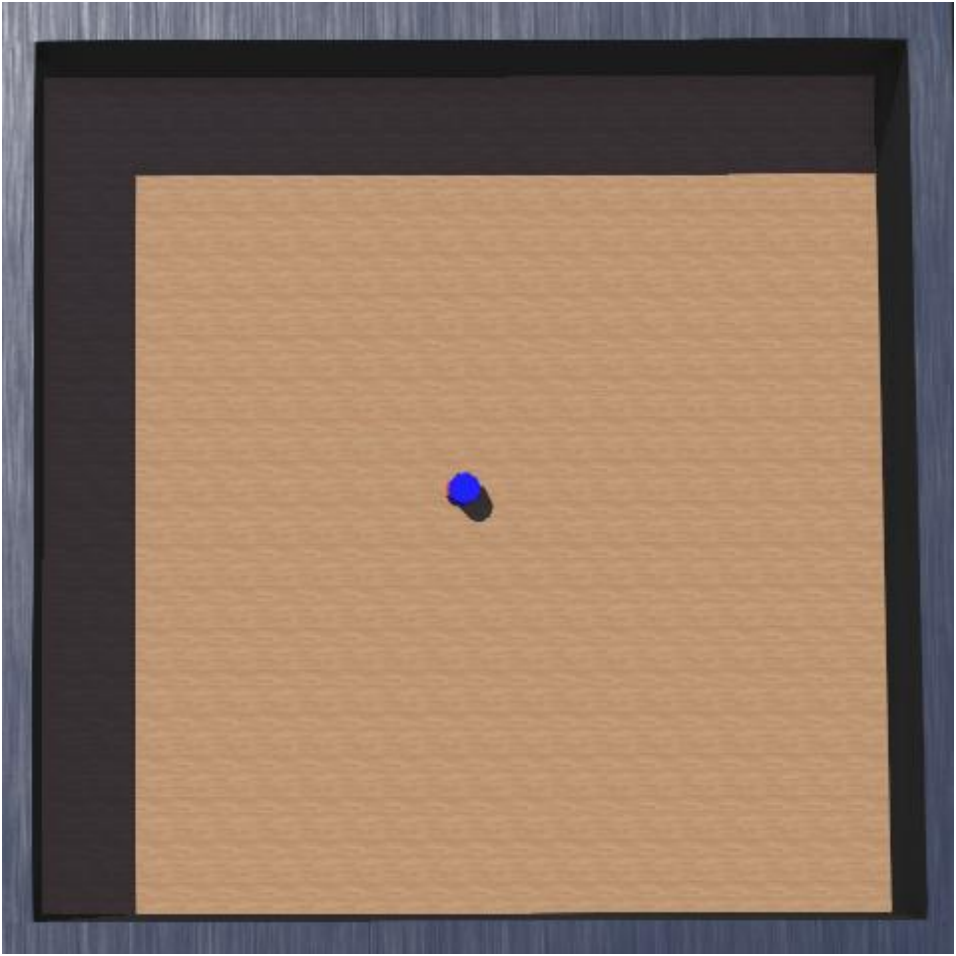

*Figure 2: The arena within which the robot operates. The robot (blue body) is shown in the arena.*

- **Sensor Characteristics:** The distance sensors have a limited range and a non-linear response curve, mimicking real-world sensor limitations. This characteristic is essential for studying how sensor limitations contribute to perceptual distortions. Specifically, the lookup table of each distance sensor maps the sensed distance to a return value. The lookup table uses a tolerance (e.g. 0.1) causing the return value to oscillate within the range defined by the tolerance.

- **Actuator Dynamics:** The robot's wheels are modeled with realistic dynamics, including inertia and friction. This ensures that the robot's movements are not instantaneous and that its actions have consequences over time. The maximum speed of each wheel is limited to a predefined value (MAX_SPEED = 2.0 m/s) to prevent unrealistic movements and ensure the stability of the simulation.

## Data Collection

The data collection process involves a random walk strategy designed to expose the robot to diverse sensorimotor experiences.

- **Random Walk Algorithm:** At each time step, the left and right wheel velocities are independently sampled from a uniform distribution between -MAX_SPEED and MAX_SPEED.

This results in unpredictable trajectories within the arena, maximizing the exploration of different positions and orientations.

- **Action Duration:** Each random action is maintained for a fixed duration (ACTION_DURATION = 5.0 seconds). This allows the robot to experience the consequences of its actions over a short period and generates a sequence of sensor readings that reflect its movement.

- **Data Logging:** The following data is recorded for each time step:
    - Left and Right Wheel Speeds: The commanded velocities for each wheel.
    - Initial and Final Positions (X, Y, Z): The robot's position before and after each action.
    - Displacement (X, Y, Z): The change in position during each action.
    - Distance Sensor Readings: The values returned by each of the sixteen distance sensors.
    - Compass Reading (Yaw): The robot's orientation.
    - Stuck Flag: A Boolean value indicating whether the robot has moved less than a threshold distance during the action duration. This flag helps identify situations where the robot might be trapped or unable to make significant progress.

This comprehensive data collection provides a rich dataset for analyzing the relationship between sensor readings, robot actions, and physical location. The periodic saving of the collected data also allows for resumption of data collection following interruptions.

## Analysis

- **Robot Path Visualization:** The collected position data is used to visualize the robot's trajectory within the arena. This provides a qualitative overview of the robot's exploration pattern.

- **K-Nearest Neighbors (KNN) Analysis:** KNN is employed to examine the relationship between sensor space and physical space. By finding the k-nearest neighbors in sensor space for a given point and then visualizing their corresponding physical locations, we can assess the degree of distortion in the perceptual space. This analysis is performed both with and without simulated sensor noise to understand the impact of noise on the mapping between perception and reality.

- **Standard Deviation and Correlation Analysis**: To assess the consistency and relationships within the sensor data we analyze standard deviations and correlations. Standard deviations of each sensor reading provide an indication of how much the sensor readings change within distinct clusters. Low standard deviations suggest consistent readings whereas high standard deviations indicate more variability, possibly due to noise or the robot's motion. Correlation analysis across different sensors provides insight into whether these sensors capture similar characteristics. High correlation coefficients between sensor readings might indicate redundant information. Comparing how standard deviations and correlations change between filtered and unfiltered data also help assess the efficacy of the filtering (on yaw, for example).

- **Convex Hull Analysis:** The convex hull of a set of points is the smallest convex polygon that encloses all the points [22]. We compute the convex hull of physically close points and their corresponding points in sensor space. This analysis reveals how the topology of spatial relationships is transformed by the robot's perception, providing insights into how the robot segments and represents its environment.

- **Clustering:** K-means clustering is applied to sensor data conditioned on the robot's yaw. This method helps identify distinct clusters in sensor space that may correspond to specific regions or features in the physical environment. The elbow method is used to determine the optimal number of clusters, ensuring that the chosen number reflects the underlying structure of the data rather than being arbitrarily imposed.

- **Transformation of Lines and Grids:** To visualize how the robot's perception distorts spatial structures, we create test lines and grids in physical space and map them to sensor space. By comparing the shape and orientation of these structures in both spaces, we can quantify the perceptual distortions introduced by the robot's sensors and movement. This is done by finding corresponding points in sensor space using a KD-tree for efficient nearest neighbor search. The median sensor values of the nearest neighbors are used to map each point from physical space to sensor space, mitigating the effect of outliers and noise.

## Results

The section will explore detailed quantitative analysis based on the methods described above. This will include statistics on KNN distances, convex hull metrics (area, perimeter), cluster characteristics (size, standard deviation of sensor values within clusters), and analysis of line/grid transformations. This rigorous quantitative analysis will complement the qualitative observations made from visualizations and strengthen the overall conclusions.

By providing this more detailed methodology and analysis, we aim to offer a more comprehensive and rigorous understanding of how perceptual distortions influence autonomous representation learning in a minimal robotic system. This enhanced approach strengthens the connection to ALife principles by demonstrating how simple agents can adapt and navigate despite the challenges posed by imperfect perception.

### 3.1 Robot Path Visualization

The robot's random walk strategy resulted in extensive exploration of the arena. The visualized path reveals a complex trajectory covering various locations and orientations. This diverse exploration is crucial for ensuring that the collected data represents a wide range of sensorimotor experiences.

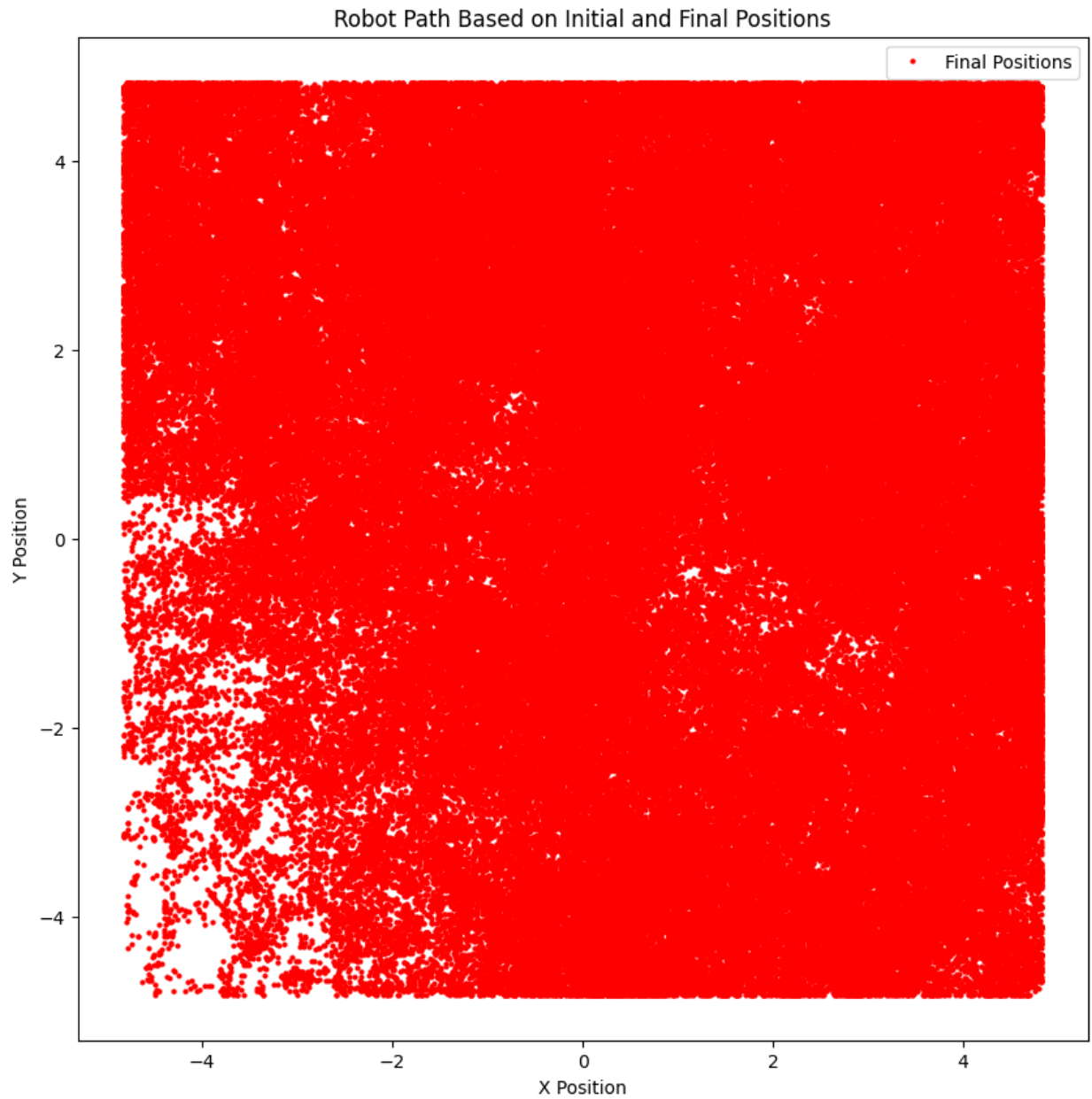

*Figure 3: Robot Path in the arena for 300,000 steps. Some regions (red) are densely explore, while others (white) are more scarcely explored.*

- **Observation:** The robot's path generally demonstrates its unbiased exploration strategy, although certain areas might be visited more frequently due to the random nature of the walk. This uneven distribution of experiences can influence the structure of the learned representation. Given enough time, exploration will cover the entire region roughly equally.

## 3.2 K-Nearest Neighbors (KNN) Analysis

KNN analysis provides insights into the relationship between sensor space and physical space. For selected points in physical space, we identified their k-nearest neighbors in sensor space. The physical distances between these neighbors were then calculated and visualized.

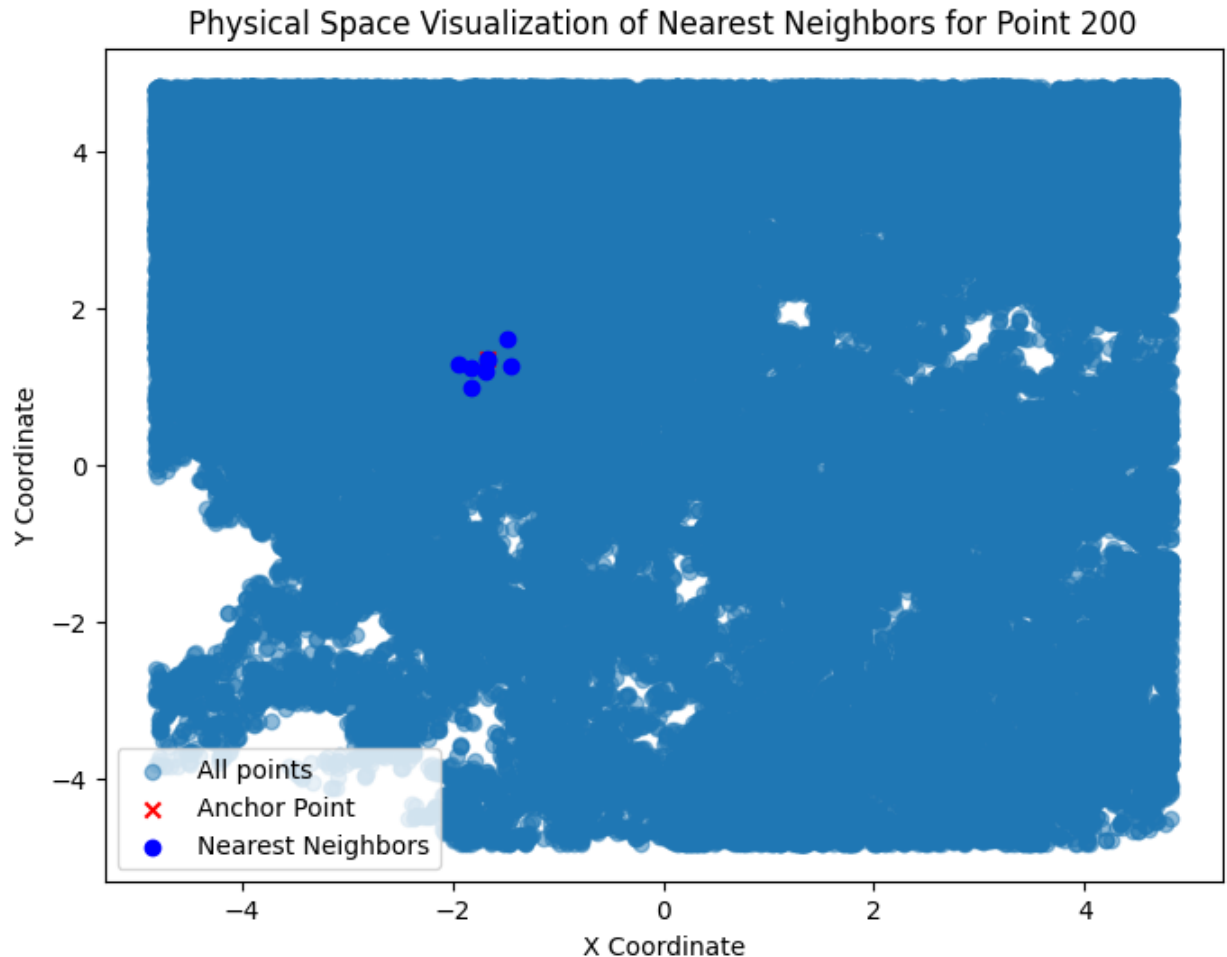

*Figure 4: Diagram showing the nearest neighbors for readings from a given point (call the anchor point). All points are shown in lighter blue while the nearest neighbors are shown in a deeper blue.*

- **Observation (Noisy Data):** With noisy sensor readings, the KNN in sensor space do not consistently correspond to physically proximate neighbors. The introduction of noise into the sensor readings causes physically distant points to cluster in sensor space while also causing physically close points to be scattered across sensor space. The physical distances to the nearest neighbors in sensor space exhibit considerable variation. This suggests that noise significantly distorts the mapping between sensor readings and physical locations.

- **Observation (Noiseless Data):** Without noise, the KNN in sensor space tend to cluster more closely in physical space. This suggests that the robot's perceptual system, in the absence of noise, can capture some degree of spatial proximity. The physical distances to the nearest neighbors exhibit less variability.

- **Implication:** Similar points seem to be near each other than far away. However, it can be seen that KNN isn't exclusive and there's some physically nearby points that aren't neighbors in the sensory space. Sensor noise plays a crucial role in shaping the robot's perceptual space. Even small amounts of noise can substantially distort the relationship between sensor readings and physical locations, making it challenging to directly infer spatial proximity from sensor data.

*3.3 Standard Deviation and Correlation Analysis*

Analysis of standard deviations and correlations of sensor readings provide further quantitative insights into the characteristics of the data.

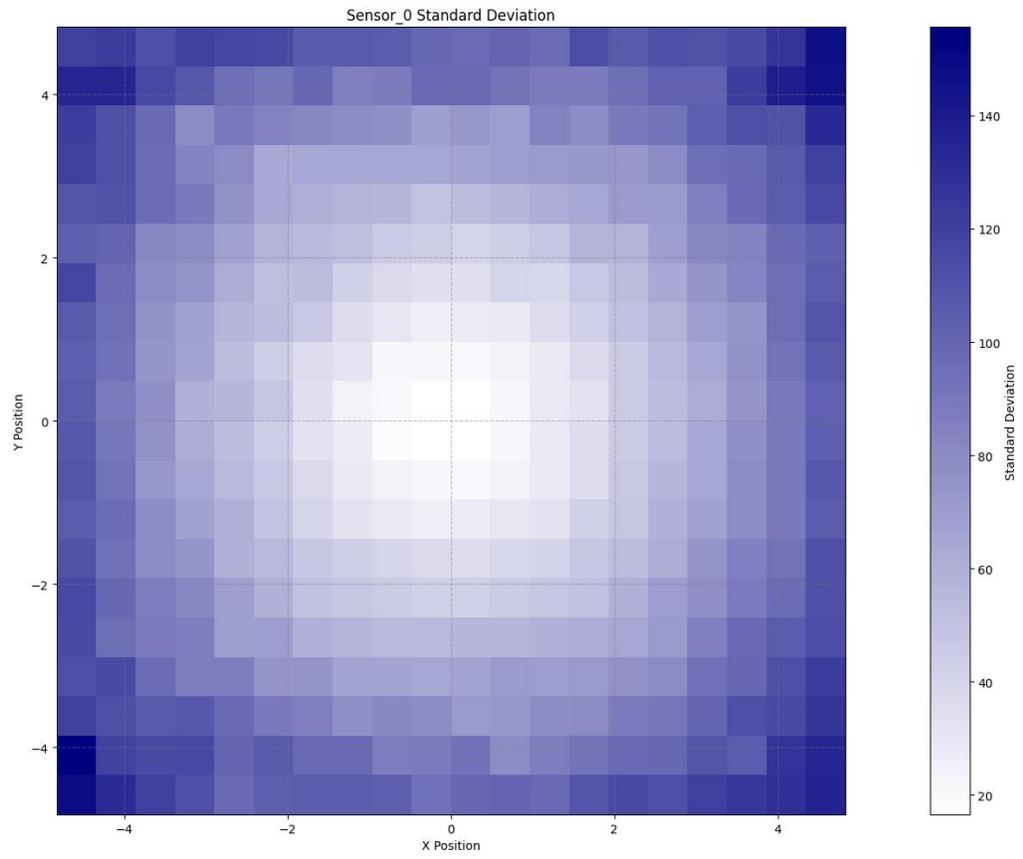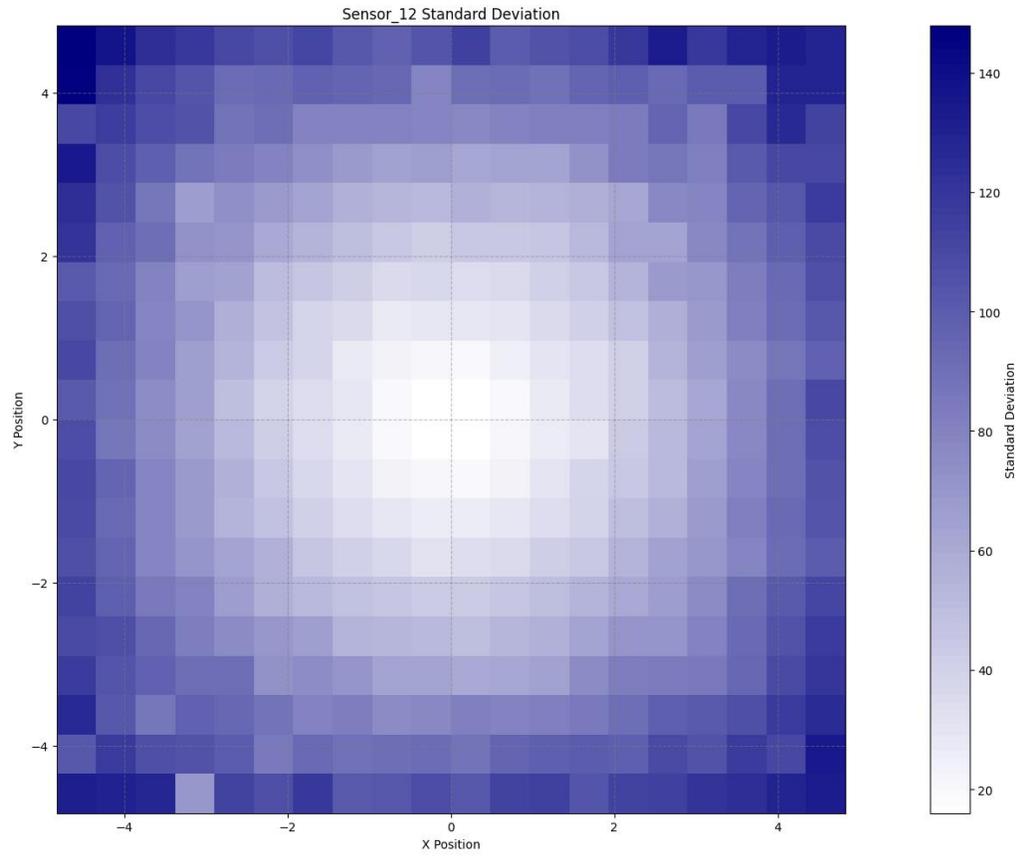

*Figure 5: Standard deviation for readings from a sensor (sensor 12). Standard deviation increases with proximity to walls (the bluer regions). This can be explained by the fact that a small angular turn leads to greater changes near the walls than away.*

- **Observation:** The standard deviations of sensor readings vary across clusters based on proximity to walls and corners. Lower standard deviations are observed when the robot is in open space compared to being close to walls or obstacles.

- **Implication:** The variability in sensor readings reflects the robot's changing environment and experiences, indicating that the perceptual space is not uniform but rather structured by the physical surroundings.

- Rolling standard deviation shows a similar pattern. To obtain rolling standard deviations, we do standard deviation on sequences of the sensor readings as they were collected in time. We use windowing. A window covers a sequence of readings. Then we average the standard deviations per grid.

- A takeaway from this is that the standard deviation and rolling standard deviations can be used to uniquely identify points within the square room. That is, they give information about where we are located. Such information is accessible through motion.

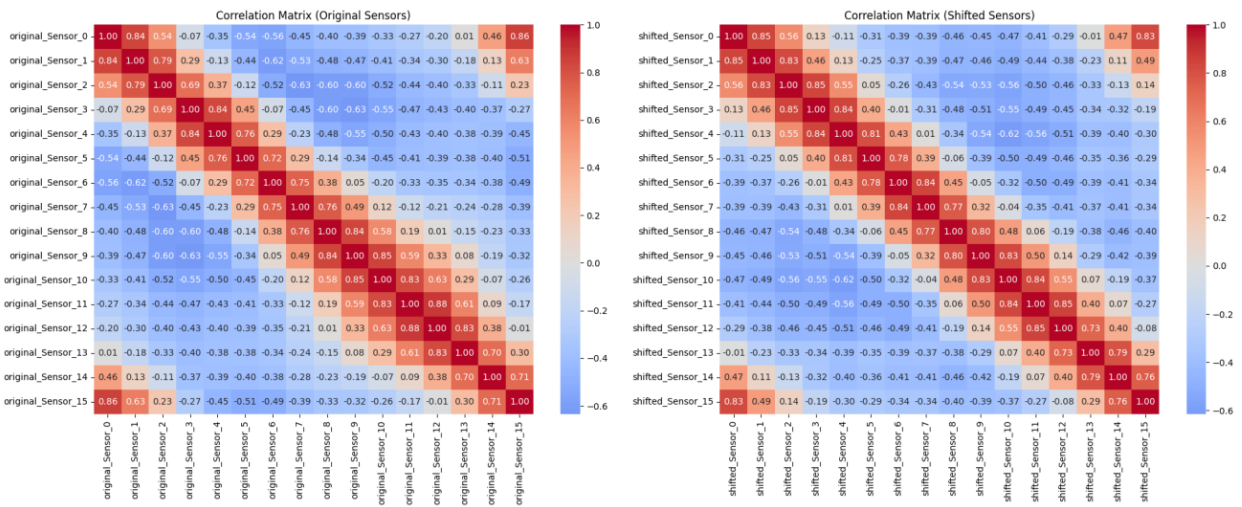

*Figure 6: Correlations between the different distance sensors. It's clear that sensors near each other have higher correlation than sensors away from each other.*

- **Observation**: There exist positive correlations in sensor readings of some adjacent sensors. Non-adjacent sensor readings show varying degrees of correlation depending on which specific sensors are considered. Filtering based on yaw causes an increase in the magnitude of correlations between sensor readings.

- **Implication**: This highlights the importance of considering the robot's orientation when analyzing sensor data, and also shows how the robot's perceptual space is influenced by its physical location, sensor noise, orientation, and more.

### 3.4 Convex Hull Analysis

We analyzed the topological correspondence between physical space and sensor space using convex hulls. By comparing the convex hull of a set of points in physical space with the convex hull of the corresponding sensor readings, we assess how spatial relationships are transformed by the robot's perception.

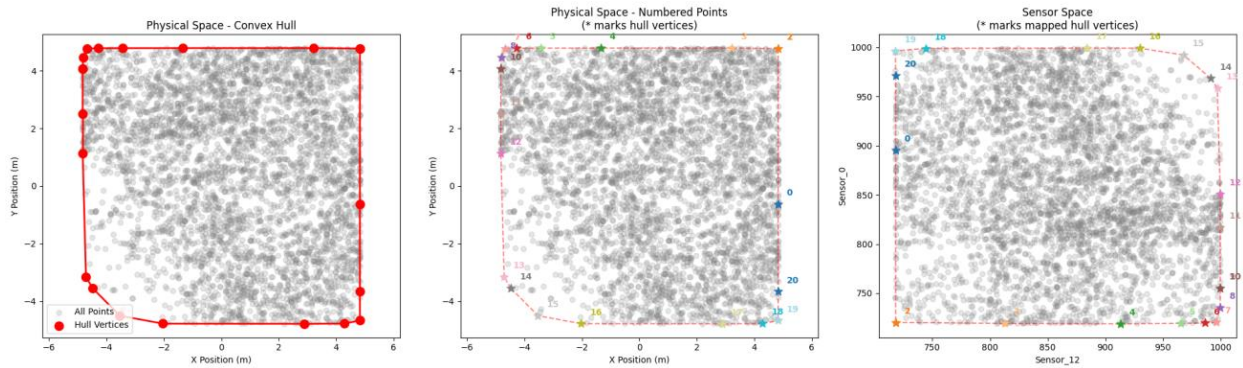

*Figure 7: Convex hull points in the physical space (left) and the same hull points in sensor space. There's a topological preservation. Sensor readings are taken within a tolerance of one of the other sensors (compass).*

- **Observation:** The shape and size of the convex hull in sensor space differ significantly from the convex hull in physical space. The transformation of the hull depends on the robot's orientation and the specific sensor readings.

- **Implication:** While the precise metric relationships are distorted in sensor space, the overall topological structure is partially preserved. This implies that the robot can still extract some information about the relative positions of objects and boundaries, even though the distances and angles might be warped.

**3.5 Clustering:**

K-means clustering was applied to sensor data conditioned on the robot's orientation (yaw) to identify emergent structures within the perceptual space. The elbow method suggested an optimal cluster number of four, corresponding to the four walls of the arena. This confirms that even with noisy and distorted sensory input, the robot is able to segment its experience based on major environmental features.

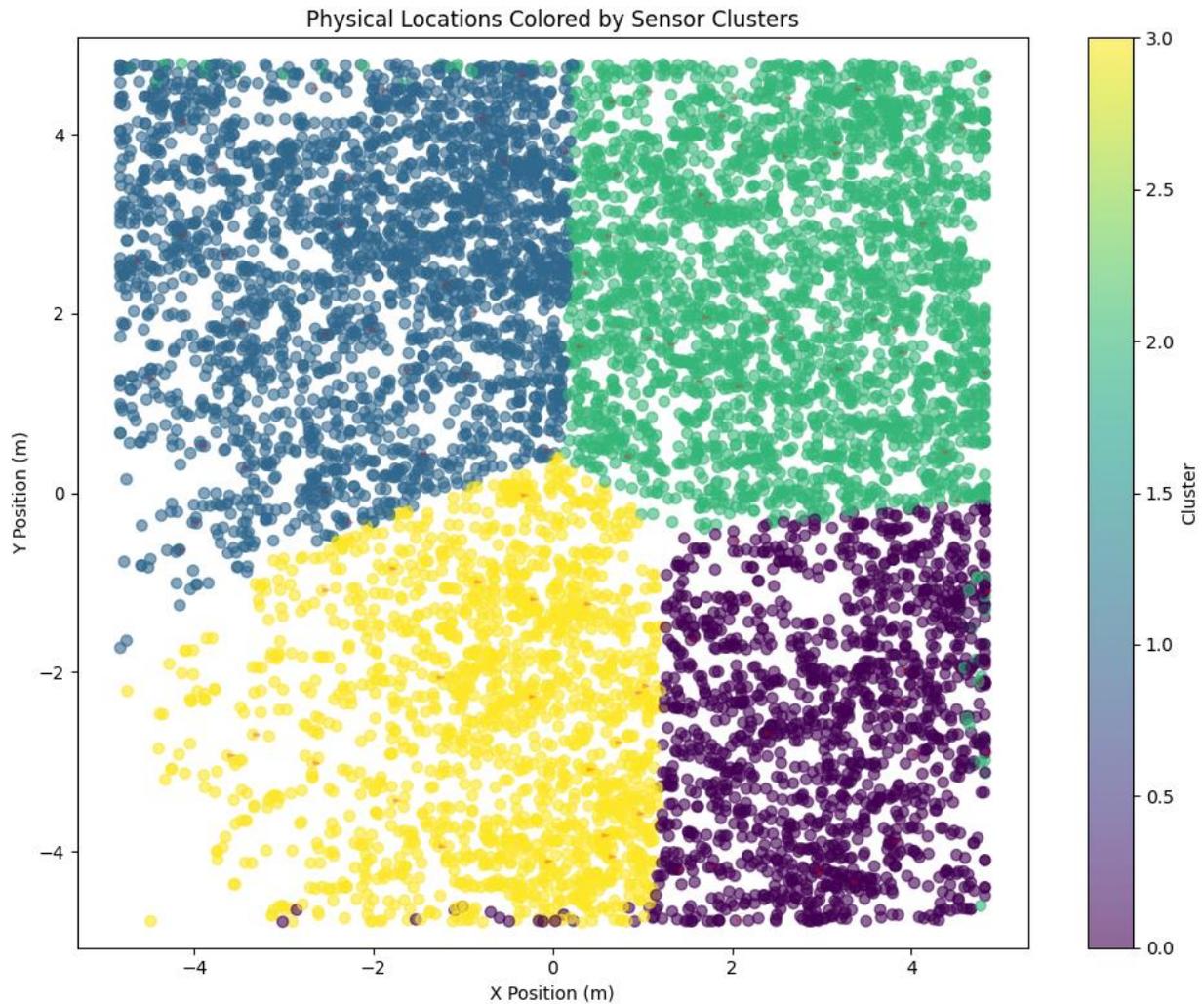

*Figure 8: Clusters for sensor readings taken within a tolerance of another sensor reading. Note that the clustering is done in sensory space and not physical space. The diagram shows a neat matching between physical space and sensor space.*

- **Observation:** The clusters in sensor space exhibit distinct patterns that correlate with the robot's proximity and orientation relative to the walls. Points belonging to the same cluster in sensor space tend to be located near the same wall in physical space.

- **Implication:** The emergence of clusters demonstrates that the robot is implicitly learning a structured representation of its environment based on sensorimotor experience. Despite the distortions, the perceptual space is organized in a way that reflects the underlying physical structure of the arena.

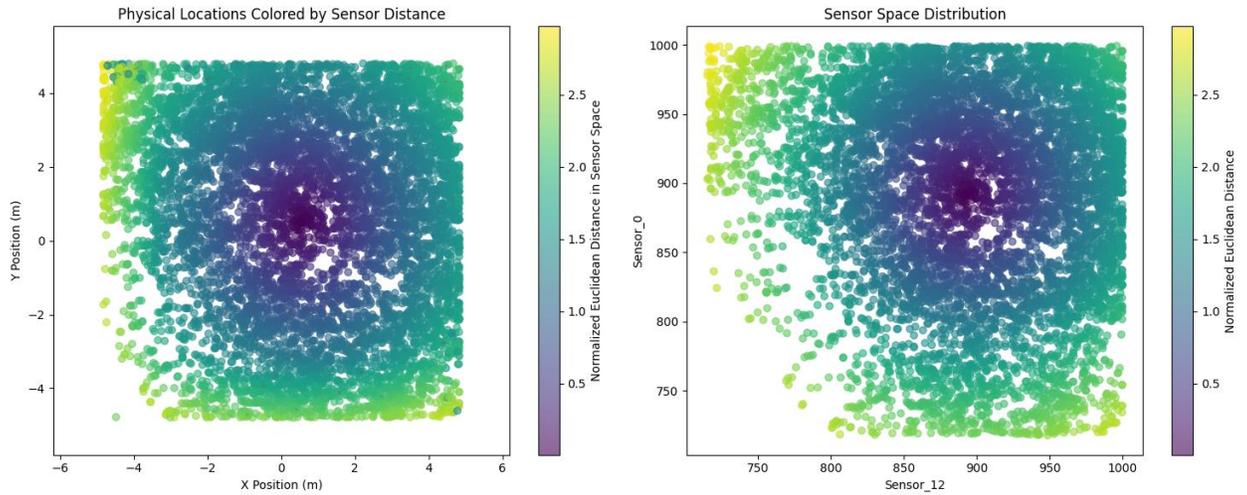

*Figure 9: Scatter plot of readings in physical space. How the readings map to sensor space.*

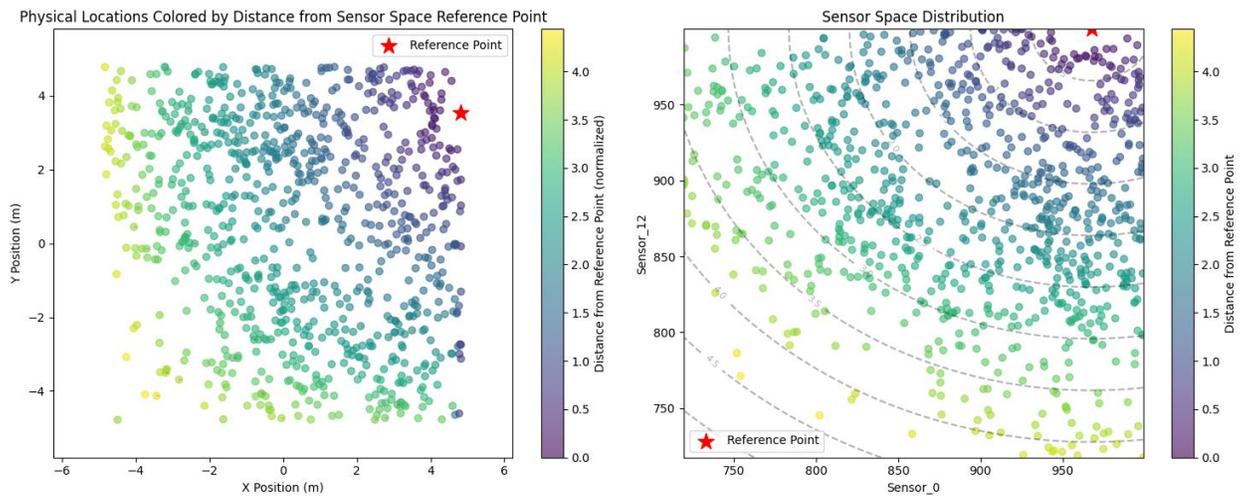

*Figure 10: Left: a sample of readings from physical location, the color showing distance from a reference point. Right: How the reference point and the rest of the other points map to sensor space.*

### 3.6 Transformation of Lines and Grids:

By mapping test lines and grids from physical space to sensor space, we visualize how the robot's perception transforms spatial structures.

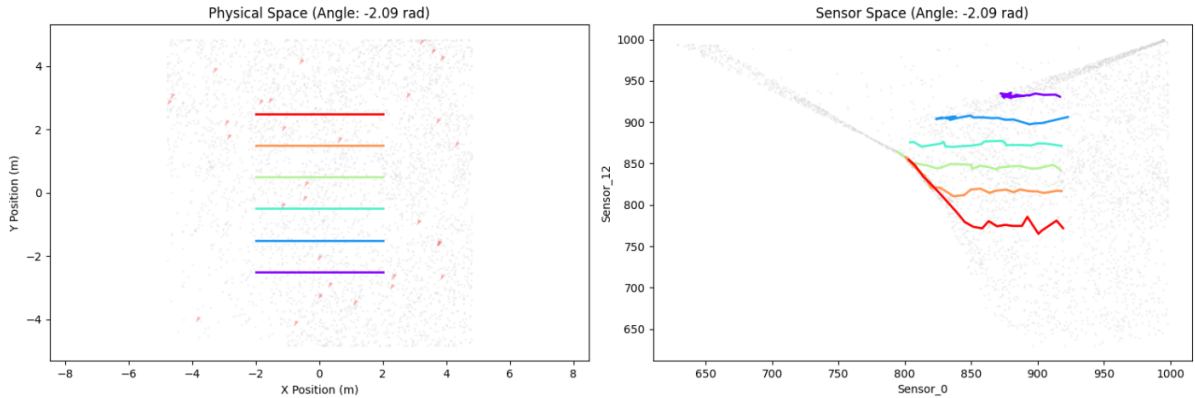

*Figure 11: The transformation of line in physical space to sensor space. Lines are distorted between physical and sensory space. The distortion is dependent on the value and tolerances for compass taken. Here we show compass readings around -2.09 radians with a tolerance of 0.01 radians.*

- **Observation (Lines):** Parallel lines in physical space are transformed into curves and distorted lines in sensor space. The degree and type of distortion depend on the robot's orientation relative to the lines.

- **Observation (Grids):** Regular grids in physical space are transformed into warped and non-uniform grids in sensor space. The grid cells become stretched and compressed, reflecting the non-linear mapping between the two spaces.

- **Implication:** These transformations highlight the complex and non-linear nature of the robot's perceptual mapping. The robot's sensorimotor experience fundamentally alters how spatial structures are represented internally, creating a distorted but functional model of the environment.

### 3.7 Implications for Embodied Cognition and Artificial Life:

The results of our experiments have broader implications for understanding embodied cognition and the development of autonomous agents in artificial life:

- **Embodied Representation Learning:** The robot's ability to learn a structured representation of its environment without explicit spatial information underscores the importance of embodiment in shaping cognition. The robot's physical interactions and sensorimotor experiences are crucial for constructing a functional, albeit distorted, model of the world.

- **Adaptive Behavior in Distorted Spaces:** The robot's successful navigation within its distorted perceptual space demonstrates that adaptive behavior can emerge even with imperfect sensory information. This suggests that biological agents, which also face limitations in their perception, may rely on similar mechanisms to navigate their environment effectively.

- **Minimal Agency and Autonomy:** Our minimal robotic system demonstrates a basic form of autonomy. The robot is not explicitly programmed to navigate its environment but instead learns to do so through exploration and interaction. This highlights the potential for simple agents to develop complex behaviors through autonomous representation learning.

- **Evolutionary Robotics:** The insights gained from this study can inform the design of evolutionary robotics experiments. By understanding how perceptual distortions affect behavior, we can develop more effective fitness functions and selection pressures for evolving robot controllers that are robust to sensory limitations.

These expanded results provide a more thorough analysis of the robot's perceptual space and its implications for autonomous representation learning. The quantitative and qualitative findings contribute to a deeper understanding of embodied cognition, minimal agency, and the role of perception in shaping behavior in artificial life.

## Discussion

This section goes into the implications of our findings, exploring their relevance to embodied cognition, autonomous representation learning, and the broader field of artificial life. It also addresses limitations of the current study and outlines potential avenues for future research.

**4.1 Embodied Cognition and the Situated Nature of Perception:**

Our results strongly support the principles of embodied cognition, highlighting the crucial role of the body and its interactions with the environment in shaping cognitive processes. The robot's perceptual space, far from being a veridical representation of the physical world, is heavily influenced by its sensorimotor experience. The distortions observed in the KNN analysis, convex hull comparisons, and grid transformations demonstrate that the robot's perception is not simply a passive reception of sensory information but rather an active construction shaped by its embodied interactions. This aligns with the situated cognition perspective [1, 2], which emphasizes that cognition is not an abstract computation performed by a disembodied brain but rather a process grounded in the agent's specific sensorimotor capabilities and its dynamic engagement with the environment.

The observed clustering of sensor data conditioned on the robot's yaw further reinforces the embodied nature of representation learning. The robot implicitly learns to associate specific sensor patterns with particular locations and orientations in the arena, effectively building a sensorimotor map of its surroundings. This map, however distorted, allows the robot to navigate effectively and demonstrates that functional representations can emerge from embodied experience even without explicit spatial information or complex mapping algorithms. This finding resonates with research on sensorimotor contingencies [16], which emphasizes the role of predictable relationships between sensory inputs and motor outputs in shaping perception and action.

**4.2 Autonomous Representation Learning and Minimal Agency:**

The robot's ability to navigate its environment successfully despite the distorted perceptual space demonstrates a form of autonomous representation learning. The robot is not provided with a pre-defined map or explicit instructions on how to navigate. Instead, it learns to associate sensor readings with actions and outcomes through its own exploratory behavior. This self-organized learning process aligns with the principles of minimal agency [9, 10], which explores how simple agents with limited cognitive resources can exhibit complex behaviors through embodied interaction.

The random walk strategy employed in our study, while seemingly simplistic, plays a crucial role in driving the learning process. By exposing the robot to a diverse range of sensorimotor experiences, the random walk facilitates the discovery of regularities and correlations within the sensor data. These regularities, in turn, form the basis for the emergent structured representation that enables navigation. This highlights the importance of exploration and diversity of experience in autonomous representation learning.

### 4.3 Implications for Artificial Life and Robotics:

Our findings have several implications for the design and development of artificial agents in both artificial life research and robotics:

- **Robustness to Sensor Noise and Imperfect Perception:** The robot's ability to navigate effectively despite noisy and distorted sensor readings suggests that robust behavior can emerge from relatively simple mechanisms. This has implications for building robots that can operate in real-world environments, where sensor noise and perceptual errors are unavoidable.

- **Simplified Mapping and Navigation:** The robot's implicit mapping strategy, based on sensorimotor experience rather than explicit map building, offers a potentially more efficient and robust approach to navigation. This could be particularly beneficial in situations where computational resources are limited or the environment is highly dynamic.

- **Understanding Biological Navigation:** The robot's behavior provides insights into how biological agents might navigate their environment using limited sensory information. The distorted perceptual space observed in our study could reflect similar distortions in the perceptual systems of animals, suggesting that biological navigation might rely on more flexible and adaptable representations than previously assumed.

- **Evolutionary Robotics:** Our findings have direct relevance to evolutionary robotics, where the goal is to evolve robot controllers for specific tasks. By understanding how perceptual distortions affect behavior, we can design more effective fitness functions and selection pressures for evolving robust and adaptive controllers.

### 4.4 Limitations and Future Research:

While our study provides valuable insights into perceptual distortions and autonomous representation learning, it also has limitations that warrant further investigation:

- **Simplified Environment:** The robot operates within a simple square arena. Future research should explore more complex environments with obstacles, varying terrain, and dynamic elements to understand how the robot's perceptual space and navigation strategies adapt to greater environmental complexity.

- **Limited Sensor Modalities:** The robot relies solely on distance sensors and a compass. Incorporating other sensor modalities, such as vision or touch, could enrich the robot's perceptual experience and lead to more sophisticated representations. Furthermore, analyzing how sensor fusion would affect the perceptual distortions is important.

- **Random Exploration Strategy:** While effective for initial exploration, the random walk strategy might not be optimal for learning complex navigation tasks. Future work could explore more

- directed exploration methods, such as curiosity-driven exploration [8] or goal-directed learning, to enhance the learning process.
- **Analysis of Learning Dynamics:** Our study focuses on the steady-state behavior of the robot. Future research could investigate the dynamics of the learning process, analyzing how the robot's perceptual space and navigation strategies evolve over time.
- **Comparison with Biological Systems:** Comparing the robot's behavior with that of biological agents, such as insects or rodents, could reveal common principles and mechanisms underlying embodied navigation and adaptation to perceptual distortions.
- **Transfer Learning and Generalization:** A key aspect of intelligent behavior is the ability to generalize learned knowledge to novel situations. Future research should investigate how the robot's learned representation can be transferred to different environments or tasks, assessing the robustness and generalizability of the autonomous learning process.
- **Higher-Dimensional Analysis:** Exploring higher-dimensional sensor spaces, by increasing the number of sensors or incorporating more complex sensor modalities, could reveal more intricate structures and relationships within the perceptual space.
- **Impact of Sensor Range and Resolution:** Investigating how the range and resolution of the distance sensors affect the degree of perceptual distortion and the quality of the learned representation is crucial for understanding the limitations and trade-offs of different sensor configurations.
- **Closed-Loop Control and Feedback:** Our current study employs open-loop control during the random walk phase. Incorporating closed-loop control and feedback mechanisms could allow the robot to actively shape its sensory input and potentially mitigate the effects of perceptual distortions.

## Conclusion

This study demonstrates how a minimal robotic system can autonomously learn a structured representation of its environment despite significant perceptual distortions. By analyzing the robot's sensor data during random exploration, we have revealed how the perceptual space is constructed, how distortions arise, and how coherent structures emerge that enable navigation. This work contributes to the field of Artificial Life by providing insights into the relationship between perception, action, and representation learning in minimal agents, with implications for understanding embodied cognition and the emergence of autonomous behavior.